\documentclass[runningheads]{llncs}
\usepackage{amsmath}
\usepackage{amsfonts}
\usepackage[dvips]{graphicx}
\graphicspath{{./eps/}}
\DeclareGraphicsExtensions{.eps}
\usepackage{algorithm}% http://ctan.org/pkg/algorithm
\usepackage{algorithmic}
\usepackage[compatibility=false]{caption}% http://ctan.org/pkg/caption
\usepackage{multirow}
\usepackage{subcaption}
\usepackage{amsmath}
\usepackage[noadjust]{cite}
\usepackage{pifont}

\begin{document}
		
\title{Improved Breast Mass Segmentation in Mammograms with Conditional Residual U-net}
\titlerunning{Breast Mass Segmentation with CRU-Net}
\author{Heyi Li\inst{1} \and
Dongdong Chen\inst{1} \and 
William H. Nailon\inst{2}  \and \\
Mike E. Davies\inst{1} \and 
David Laurenson\inst{1}}
\institute{Institute for Digital Communications,
	University of Edinburgh, Edinburgh, UK
	\{heyi.li, d.chen, dave.laurenson, mike.davies\}@ed.ac.uk
	\and Oncology Physics Department, Edinburgh Cancer Centre, Western General
	Hospital, Edinburgh, 
	bill.nailon@luht.scot.nhs.uk}
\maketitle               % typeset the header of the contribution

\begin{abstract}
We explore the use of deep learning for breast mass segmentation in mammograms. By integrating the merits of residual learning and probabilistic graphical modelling with standard U-Net, we propose a new deep network, Conditional Residual U-Net (CRU-Net), to improve the U-Net segmentation performance. Benefiting from the advantage of probabilistic graphical modelling in the pixel-level labelling, and the structure insights of a deep residual network in the feature extraction, the CRU-Net provides excellent mass segmentation performance. Evaluations based on INbreast and DDSM-BCRP datasets demonstrate that the CRU-Net achieves the best mass segmentation performance compared to the state-of-art methodologies. Moreover, neither tedious pre-processing nor post-processing techniques are not required in our algorithm.     
\keywords{Mammogram mass segmentation \and structured prediction \and deep residual learning.}
\end{abstract}
\section{Introduction}
Breast cancer is the most frequently diagnosed cancer among women across the globe. Among all types of breast abnormalities, breast masses are the most common but also the most challenging to detect and segment, due to variations in their size and shape and low signal-to-noise ratio \cite{dhungel2015deep}. An irregular or spiculated margin is the most important feature in indicating a cancer. The more irregular the shape of a mass, the more likely the lesion is malignant \cite{oliver2010review}. Oliver \textit{et al.} demonstrated in their review paper that mass segmentation provides detailed morphological features with precise outlines of masses, and plays a crucial role in a subsequent cancerous classification task \cite{oliver2010review}.

The main roadblock faced by mass segmentation algorithms is the insufficient volume of contour delineated data, which directly leads to inadequate accuracy \cite{carneiro2017review}. The U-Net \cite{ronneberger2015u}, as a Convolutional Neural Network (CNN) based segmentation algorithm, is shown to perform well with limited training data by interlacing multi-resolution information. However, the CNN segmentation algorithms including the U-Net are limited by the weak consistency of predicted pixel labels over homogeneous regions. To improve the labelling consistency and completeness, probabilistic graphical models \cite{chen2017graph} have been applied for mass segmentation, including Structured Support Vector Machine (SSVM) \cite{dhungel2015deep1} and Conditional Random Field (CRF) \cite{dhungel2015deep} as a post-processing technique. To train the CRF integrated network in an end-to-end way, the CRF with the mean-field inference is realised as a recurrent neural network \cite{zheng2015conditional}. This is applied on mass segmentation \cite{zhu2017adversarial}, and achieved the state-of-art mass segmentation performance. Another limitation of CNN segmentation algorithms is that as the depth of the CNNs increase for better performing deep features, they may suffer from the gradient vanishing and exploding problems, which are likely to hinder the convergence \cite{he2016deep}. Deep residual learning is shown to address this issue by mapping layers with residuals explicitly instead of mapping the deep network directly \cite{he2016deep}.      
\begin{figure}[t!]
	\normalsize
	\centering
	\includegraphics[width=1\textwidth]{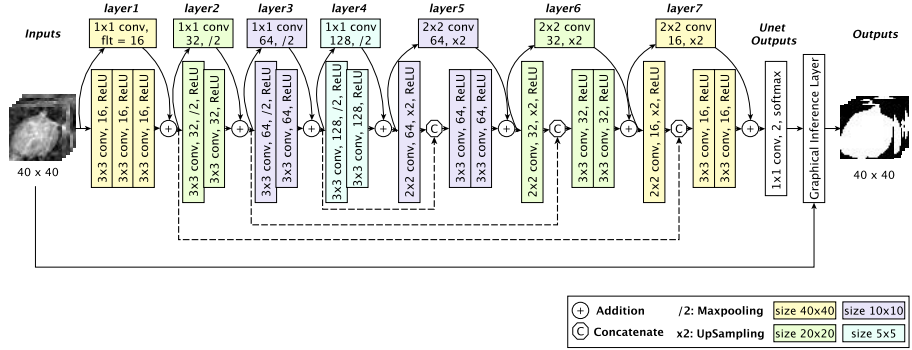}
	\caption{Proposed CRU-Net Structure}
	\label{res_unet}
\end{figure}

In this work, the CRU-Net is proposed to precisely segment breast masses with small-sample-sized mammographic datasets. Our main contributions include: 1) the first neural network based segmentation algorithm that considers both pixel-level labelling consistency and efficient training via integrating the U-Net with CRF and deep residual learning; 2) the first deep learning mass segmentation algorithm, which does not require any pre-processing or post-processing techniques; 3) the CRU-Net achieves the best mass segmentation performance on the two most commonly used mammographic datasets when compared to other related methodologies.

\section{Methodology}
The proposed algorithm CRU-Net is schematically shown in Fig. \ref{res_unet}. The inputs are mammogram regions of interest (ROIs) that contain masses and the outputs are the predicted binary images. In this section, a detailed description of applied methods is introduced: our U-Net with residual learning, followed by the pixel-level labelling with graphical inference.  

\subsection{U-Net with Residual Learning} 
The U-Net is shown to perform well with a limited volume of training data for segmentation problems in medical imaging \cite{ronneberger2015u}, which suits our situation. However, the gradient vanishing and explosion problem, which hinders the convergence, is not considered in the U-Net. We integrate residual learning into the U-Net to precisely segment breast masses over a small sample size training data. Assuming $\boldsymbol{x}: \mathrm{\Omega}\rightarrow\mathbb{R}$ ($\mathrm{\Omega}$ represents the image lattice) as an ROI and $\boldsymbol{y}: \mathrm{\Omega}\rightarrow\{0,1\}$ as the corresponding binary labelling image (0 denotes background pixels and 1 for the mass pixels), the training set can be represented by $\mathcal{D}=\{(\boldsymbol{x}^{(n)},\boldsymbol{y}^{(n)})\}_{n\in\{1,\dots,N\}}$.

The U-Net comprises of a contractive downsampling and expansive upsampling path with skip connections between the two parts, which makes use of standard convolutional layers. The output of $m$th layer with input $\boldsymbol{x}^{(n)}$ at pixel $(i,j)$ is formulated as follows:
 
\begin{equation}
{y}^{(n,m)}_{i,j} = h_{ks}(\{\boldsymbol{x}_{s_i+\delta_i, s_j+\delta_j}\}_{0\leq\delta_i,\delta_j\leq k})
\label{layer_func}
\end{equation}
where $k$ represents for kernel size, $s$ for stride or maxpooling factor, and $h_{ks}$ is the layer operator including convolution, maxpooling and the ReLU activation function. 

\begin{figure}[!t]
	\normalsize
	\centering
	\includegraphics[width=0.95\textwidth]{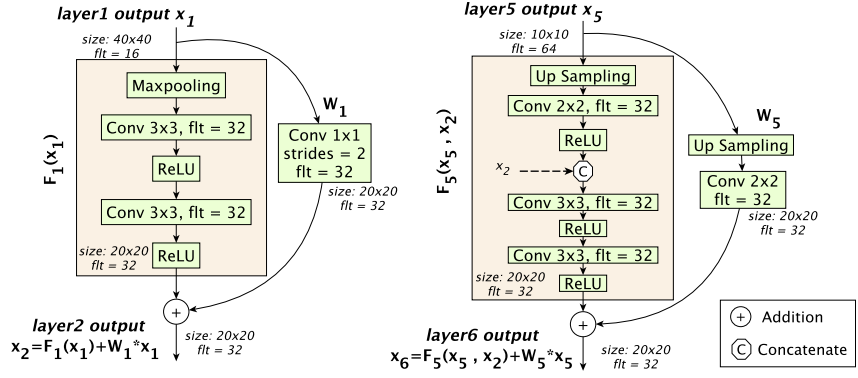}
	\caption{Residual Learning illustration for layer2 and layer6. Other layers are equivalent to this example but with different parameters.}
	\label{res}
\end{figure}

Then we integrate the residual learning into the U-Net, which solves the applied U-Net network mapping  $\mathcal{H}(\boldsymbol{x})$ with:
\begin{equation}
\mathcal{F}(\boldsymbol{x}):=\mathcal{H}(\boldsymbol{x})-W*\boldsymbol{x}
\end{equation} 
thus casting the original mapping into $\mathcal{F}(\boldsymbol{x})+W*\boldsymbol{x}$, where $W$ is a convolution kernel and linearly projects $x$ to match $\mathcal{F}(\boldsymbol{x})$'s dimensions as Fig. \ref{res_unet}. As the U-Net layers resize the image, residuals are linearly projected either with $1\times 1$ kernel convolutional layer along with maxpooling or upsampling and $2 \times 2$ convolution to match dimensions. The detailed residual connections of layer 2 and layer 6 are described in Fig. \ref{res}. These layers are shown as examples as all residual layers have analogous structure. In the final stage, a $1 \times 1$ convolutional layer with softmax activation creates a pixel-wise probabilistic map of two classes (background and masses). The residual U-Net loss energy for each output during training is defined with categorical cross-entropy. Mathematically,
\begin{equation}
f=-\sum_{i,j}\log P\big({y}_{i,j}^{(n)}\mid\boldsymbol{x}^{(n)};\boldsymbol{\theta}\big)
\label{loss_unet}
\end{equation}
where $P$ is the residual U-Net output probability distribution at position $(i,j)$ given the input ROI $\boldsymbol{x}^{(n)}$ and parameters $\boldsymbol{\theta}$.

Note that the standard U-Net is designed for images of size $572 \times 572$. Here we modify the standard U-Net to adapt mammographic ROIs ($40 \times 40$) with zero-padding for downsampling and upsampling. Residual short-cut additions are calculated in each layer. Afterthat, feature maps are concatenated as: layer 1 with layer 7, layer 2 with layer 6, layer 3 with layer 5 as shown in Fig. \ref{res_unet}. Both original ROIs and U-Net Outputs are then fed into the graphical inference layer.

\subsection{Graphical Inference}
Graphical models are recently applied on mammograms for mass segmentation. Among them, CRF incorporates the label consistency with similar pixels and provide sharp boundary and fine-grained segmentation. Mean field iterations are applied as the inference method to realise the CRF as a stack of RNN layers \cite{zheng2015conditional, zhu2017adversarial}. The cost function for CRF ($g$) can be defined as follows:
\begin{equation}
g = A(\boldsymbol{x}^{(n)}) - \exp\big(\sum_{i,j\in V} P\big({y}_{i,j}^{(n)}) + \sum_{p,q\in{E}}\phi(y_{p}^{(n)},y_q^{(n)} \mid\boldsymbol{x}^{(n)})\big)
\label{loss_crf}
\end{equation}
where A is the partition function, $P$ is the unary function which is calculated on the residual U-Net output, and $\phi$ is the pair-wise potential function which is defined with the label compatibility $\mu(y_p^{(n)},y_q^{(n)})$ for position $p$ and $q$ \cite{zheng2015conditional}, Gaussian kernels $k_G^1$, $k_G^2$ and corresponding weights $\omega_G^{(1)}$, $\omega_G^{(2)}$ \cite{pairwise_Gaussian} as $\phi(y_p^{(n)}, y_q^{(n)} \mid\boldsymbol{x}^{(n)}) = \mu(y_p^{(n)}, y_q^{(n)})\big(\omega_G^{(1)}k_G^{(1)}(\boldsymbol{x}^{(n)})+\omega_G^{(2)}k_G^{(2)}(\boldsymbol{x}^{(n)})\big)$ \cite{dhungel2015deep,zhu2017adversarial}.

Finally, by integrating (\ref{loss_unet}) and (\ref{loss_crf}) the total loss energy in the CRU-Net for each input $\boldsymbol{x}^{(n)}$ is defined as:
\begin{equation}
\ell = (1-\lambda)f + \lambda \cdot g(f,\boldsymbol{x}^{(n)})\\
\label{loss}
\end{equation}
where $\lambda \in [0,1]$ is a trade-off factor, which is empirically chosen as 0.67. And the whole CRU-Net is trained by backpropagation.

\section{Experiments}
\subsection{Datasets}
The proposed method is evaluated on two publicly available datasets INbreast \cite{moreira2012inbreast} and DDSM-BCRP \cite{heath2000digital}. INBreast is a full-field digital mammographic dataset (70$\mu m$ pixel resolution), which is annotated by a specialist with lesion type and detailed contours for each mass. 116 accurately annotated masses are contained with mass size ranging from 15$mm^{2}$ to 3689$mm^{2}$. The DDSM-BCRP \cite{heath2000digital} database is selected from the Digital Database for Screening Mammography (DDSM) database, which contains digitized film screen mammograms (43.5 microns resolution) with corresponding pixel-wise ground truth provided by radiologists. 

To compare the proposed methods with other related algorithms, we use the same dataset division and ROIs extraction as \cite{dhungel2015deep, dhungel2015deep1,zhu2017adversarial}, in which ROIs are manually located and extracted with rectangular bounding boxes and then resized into $40 \times 40$ pixels using bicubic interpolation \cite{dhungel2015deep}. In work \cite{dhungel2015deep,dhungel2015deep1,zhu2017adversarial}, extracted ROIs are pre-processed with the Ball and Bruce technique \cite{ball2007digital}, which our algorithms do not require. The INbreast dataset is divided into 58 training and 58 test ROIs; The DDSM-BCRP is divided into 87 training and 87 test ROIs \cite{dhungel2015deep}. The training data is augmented by  horizontal flip, vertical flip, and both horizontal and vertical flip. 
\begin{table}[!t]
	\renewcommand{\arraystretch}{1}
	\caption{Mass segmentation performance (DI, \%) of the CRU-Net and several state-of-te-art methods on test sets. $\lambda$ is the trade off loss factor as (\ref{loss}).}
	\label{table_results}
	\centering
	\begin{tabular}{c|c|c|c|c|c}
		\hline
		\bfseries Methodology & \mdseries INbreast & \mdseries DDSM-BCRP & \mdseries Residual & \mdseries Preprocess & \mdseries Postprocess \\ 
		\hline 
		Cardoso \textit{et. al.} \cite{cardoso2015closed} & $88$ & - & - & - &-\\
		
		Beller \textit{et. al.} \cite{beller2005example} & - & $70$ & - & - & -\\
		 
		Dhungel \textit{et. al.} \cite{dhungel2015deep1} & $88$ & $87$ & $\times$ & \checkmark & \checkmark\\
		 
		Dhungel \textit{et. al.} \cite{dhungel2015deep} & $90$ & $90$ & $\times$ & \checkmark & \checkmark\\
		
		Zhu \textit{et. al.} \cite{zhu2017adversarial} & $89.36\pm0.37$ & $90.62\pm0.16$ & $\times$ & \checkmark &$\times$\\
		\hline 
		U-Net & $92.99\pm0.23$ & $90.08\pm0.62$ & $\times$ & $\times$ & $\times$\\
		\hline
		
		%92.23
		CRU-Net ($\lambda=0$) & $92.72\pm0.09$ & $\mathbf{91.43}\pm0.02$ & \checkmark & $\times$ & $\times$\\
		CRU-Net ($\lambda=1$)  & ${92.60}\pm0.24$ & $91.41\pm0.02$ & \checkmark & $\times$ & $\times$\\
		CRU-Net, No R ($\lambda=0.67$) & $\mathbf{93.66}\pm0.10$ & $91.14\pm0.09$ & $\times$ & $\times$ & $\times$\\
		CRU-Net ($\lambda=0.67$)  & ${93.32}\pm0.12$ & ${90.95\pm0.26}$ & \checkmark & $\times$ & $\times$
	\end{tabular}
\end{table}   
\subsection{Experiment Configurations}
In this paper, each component of the CRU-Net is experimented, including $\lambda=0,1,0.67$ and the CRU-Net without residual learning (CRU-Net, No R). In the CRU-Net, convolutions are first computed with kernel size $3 \times 3$, which are then followed by a skip to compute the residual as shown in Fig. \ref{res_unet}. The feature maps in each downsampling layer are of size 16, 32, 64, and 128 respectively, while the ROIs spatial dimensions are $40 \times 40$, $20 \times 20$, $10 \times 10$ and $5 \times 5$. To avoid over-fitting, dropout layers are involved with 50\% dropout rate. The resolution of two datasets are different, with the DDSM's much higher than the INbreast's. To address this, the convolutional kernel size for DDSM is chosen as $7 \times 7$ by experimental grid search. All other hyper parameters are identical. The whole CRU-Net is optimized by the Stochastic Gradient Descent algorithm with the Adam update rule. 

\begin{figure}[!h!]
	\normalsize
	\centering	
	\begin{subfigure}[b]{0.495\linewidth}
		\includegraphics[width=\linewidth]{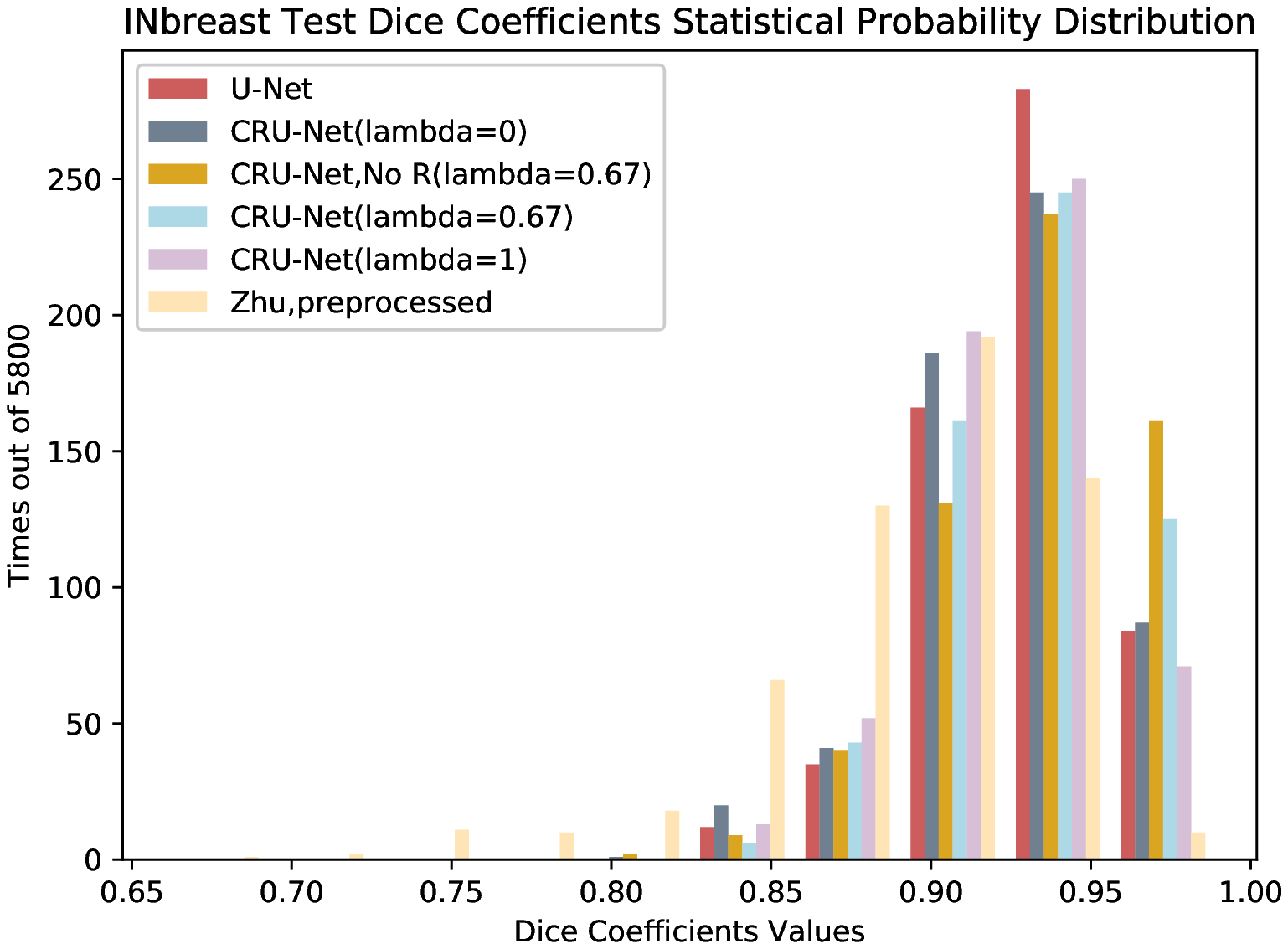}
	\end{subfigure}
	\begin{subfigure}[b]{0.495\linewidth}
		\includegraphics[width=\linewidth]{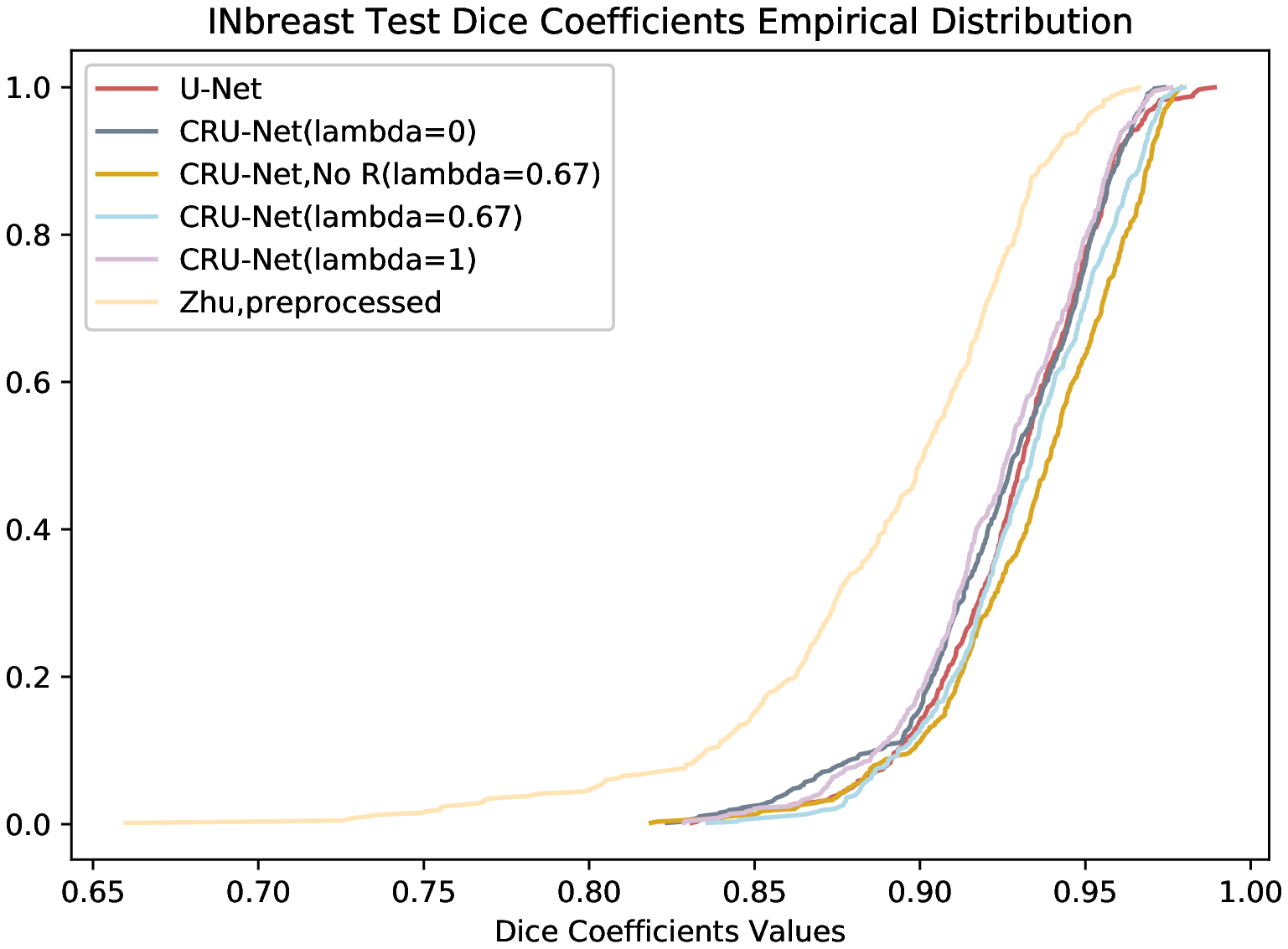}
	\end{subfigure}
	\begin{subfigure}[b]{0.495\linewidth}
		\includegraphics[width=\linewidth]{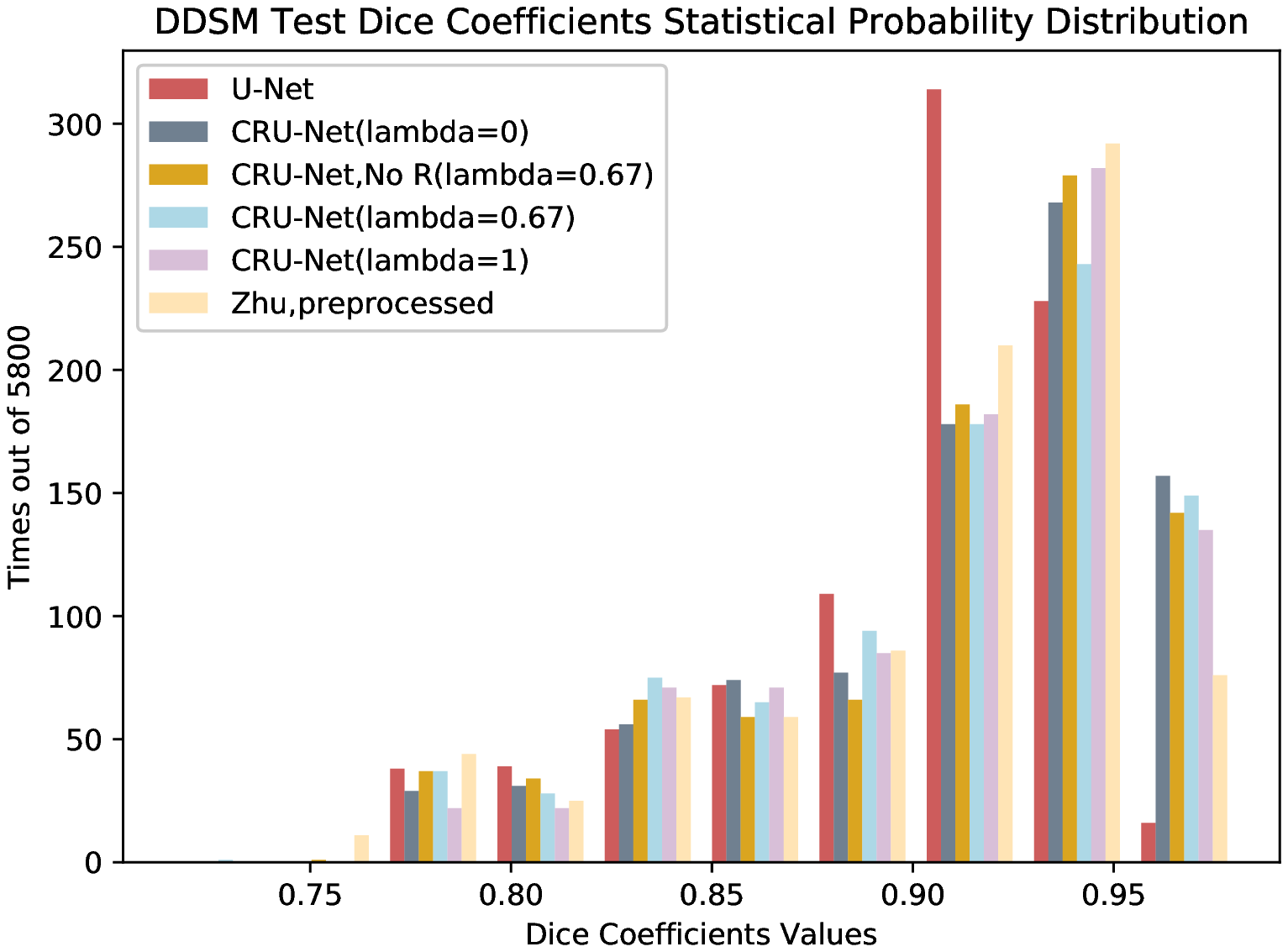}
	\end{subfigure}
	\begin{subfigure}[b]{0.495\linewidth}
		\includegraphics[width=\linewidth]{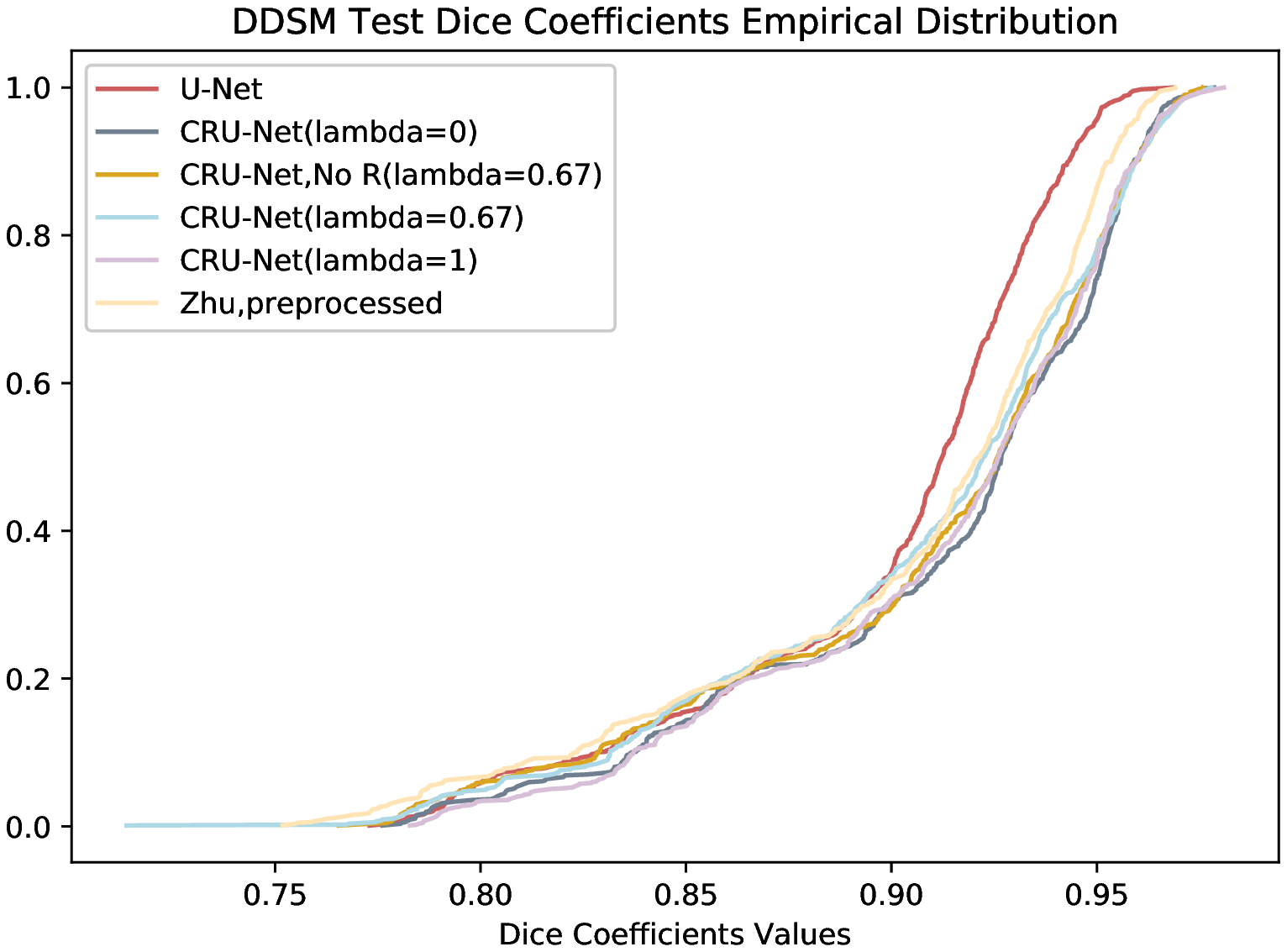}
	\end{subfigure}	
	\caption{Test Dice Coefficients Distribution of INbreast Dataset. The first row shows the distribution of INbreast dataset and the second row shows DDSM's. The left figures depict the histogram of test dice coefficients and the rights show the sampled cumulative distribution.}
	\label{dice_pdfs}
\end{figure}

\subsection{Performance and Discussion}
\begin{figure}[h!]
	\normalsize
	\centering
	\begin{subfigure}[b]{0.158\linewidth}
		\includegraphics[width=\linewidth]{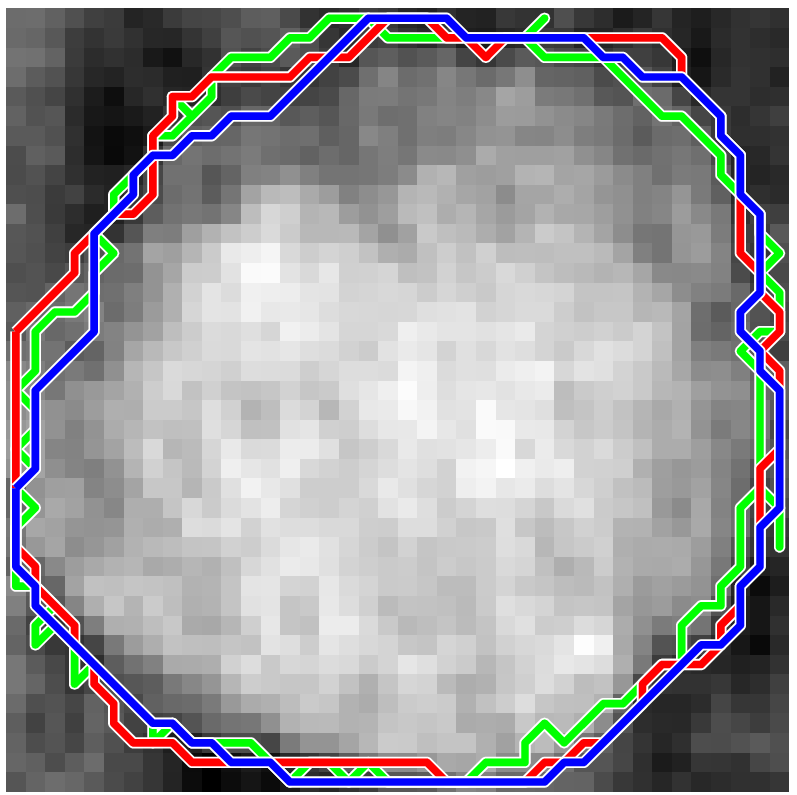}
		\caption{}
	\end{subfigure}
	\begin{subfigure}[b]{0.158\linewidth}
		\includegraphics[width=\linewidth]{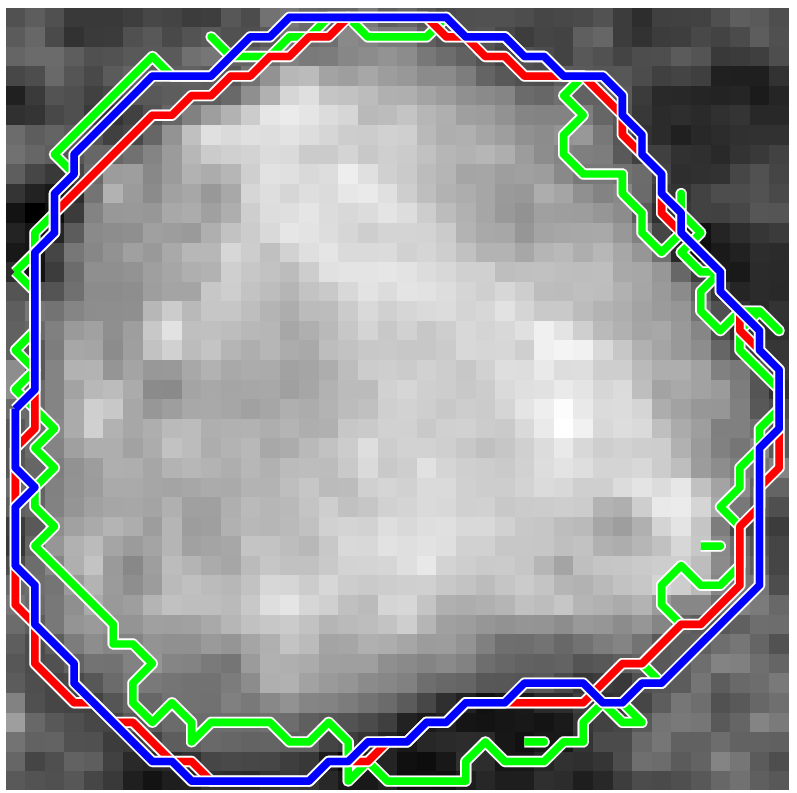}
		\caption{}
	\end{subfigure}
	\begin{subfigure}[b]{0.158\linewidth}
		\includegraphics[width=\linewidth]{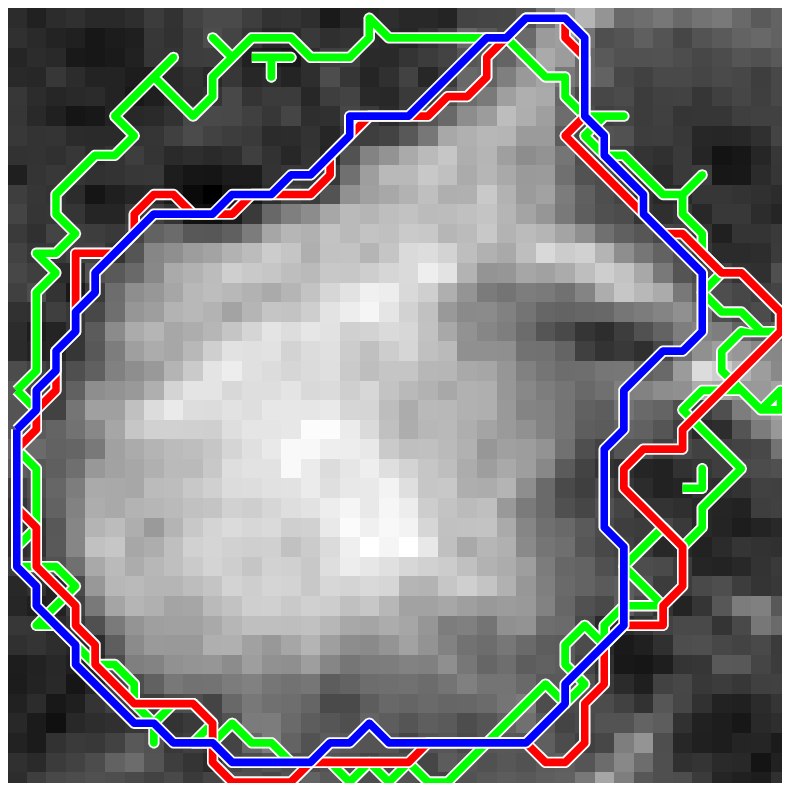}
		\caption{}
	\end{subfigure}
	\begin{subfigure}[b]{0.158\linewidth}
		\includegraphics[width=\linewidth]{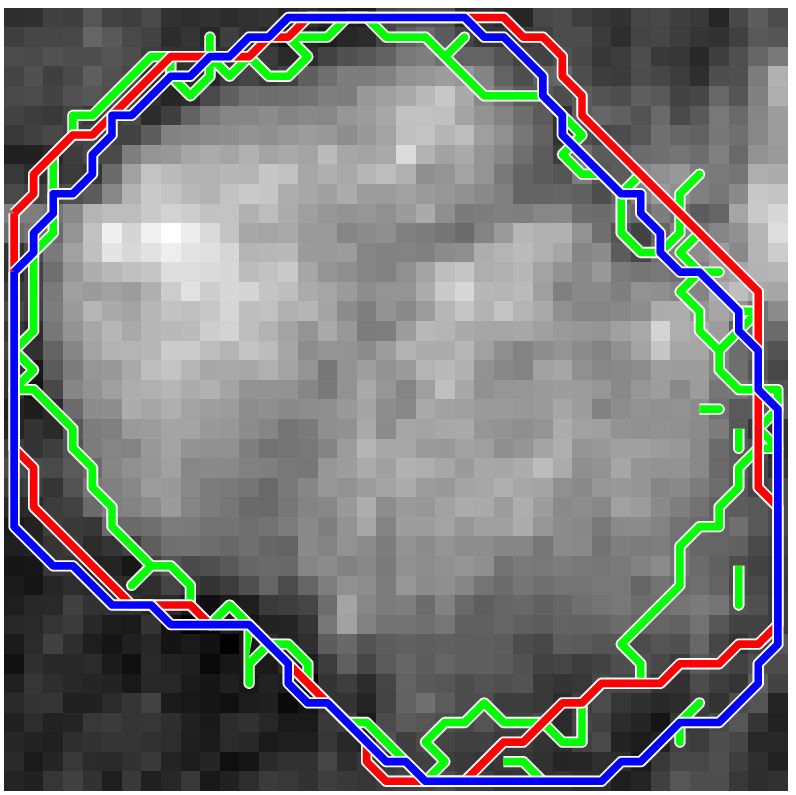}
		\caption{}
	\end{subfigure}
	\begin{subfigure}[b]{0.158\linewidth}
		\includegraphics[width=\linewidth]{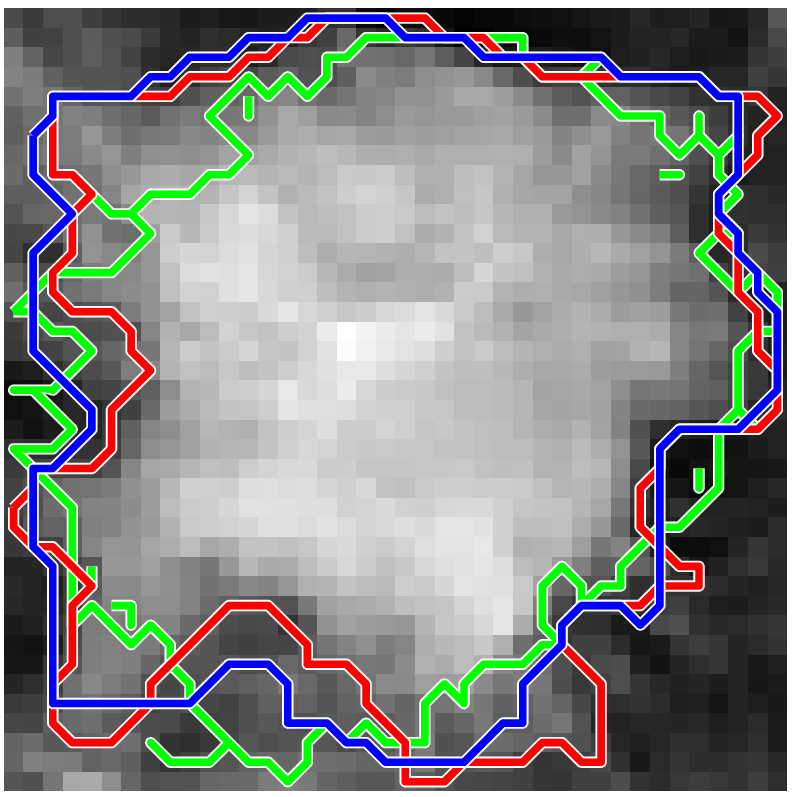}
		\caption{}
	\end{subfigure}
	\begin{subfigure}[b]{0.158\linewidth}
		\includegraphics[width=\linewidth]{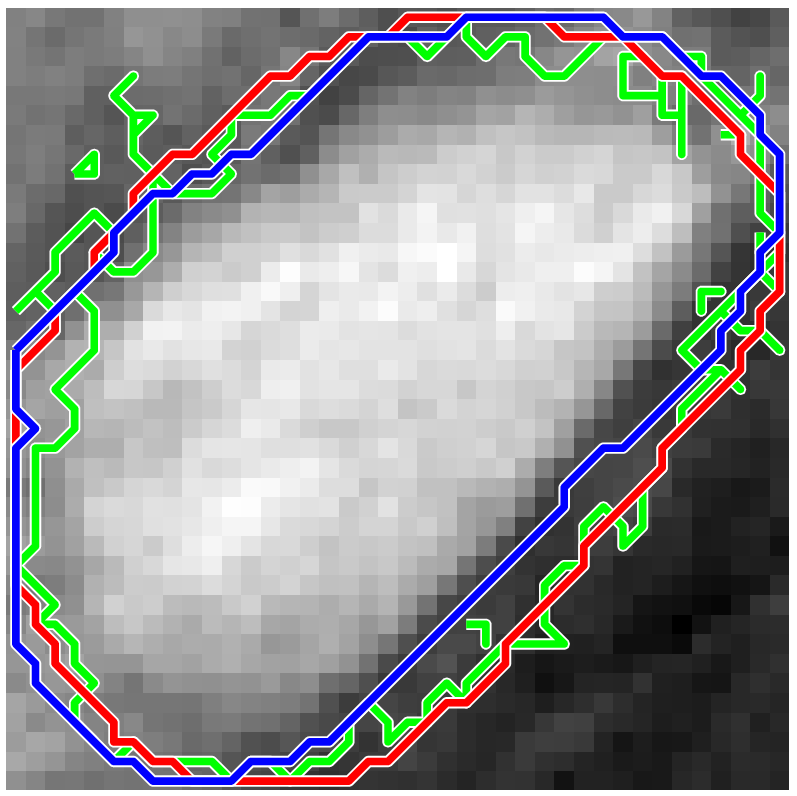}
		\caption{}
	\end{subfigure}
	\caption{Visualized comparison of segmentation results (DI $>$ 81\%) between CRU-Net and Zhu's work. Red lines denote the radiologist's contour, blue lines are the CRU-Net's results ($\lambda=0.67$), and green lines denote Zhu's method results.}	
	\label{results_contour}
\end{figure}

\begin{figure}[h!]
	\normalsize
	\centering
	\begin{subfigure}[b]{0.158\linewidth}
		\includegraphics[width=\linewidth]{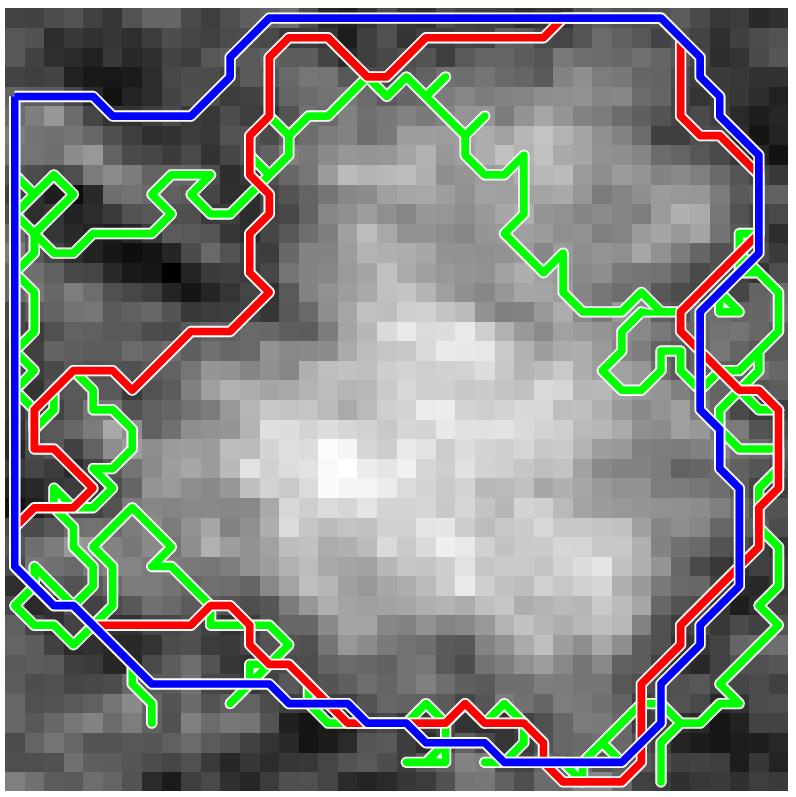}
		\caption{}
	\end{subfigure}
	\begin{subfigure}[b]{0.158\linewidth}
		\includegraphics[width=\linewidth]{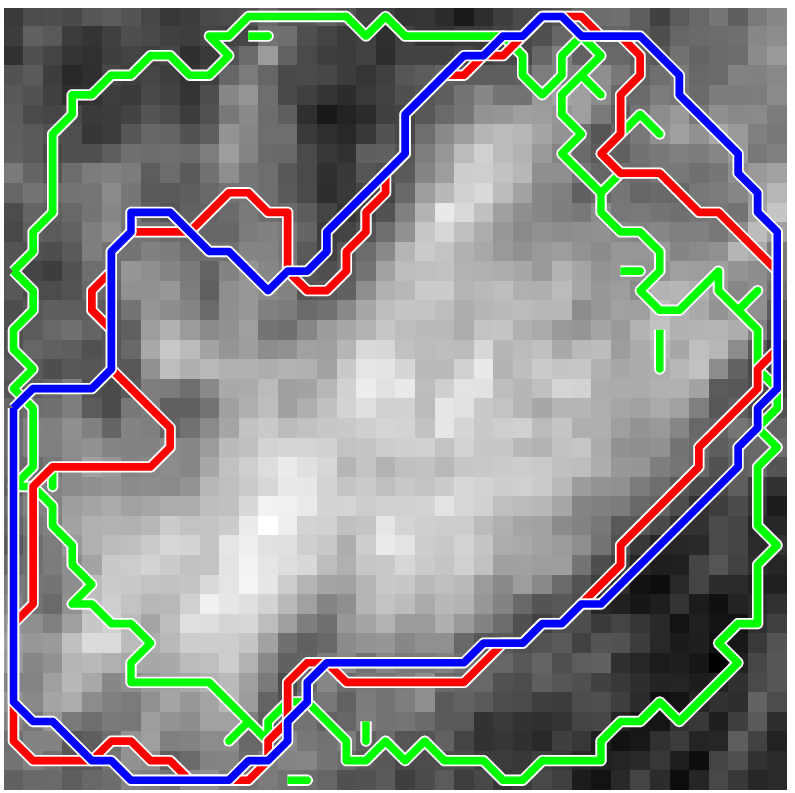}
		\caption{}
	\end{subfigure}
	\begin{subfigure}[b]{0.158\linewidth}
		\includegraphics[width=\linewidth]{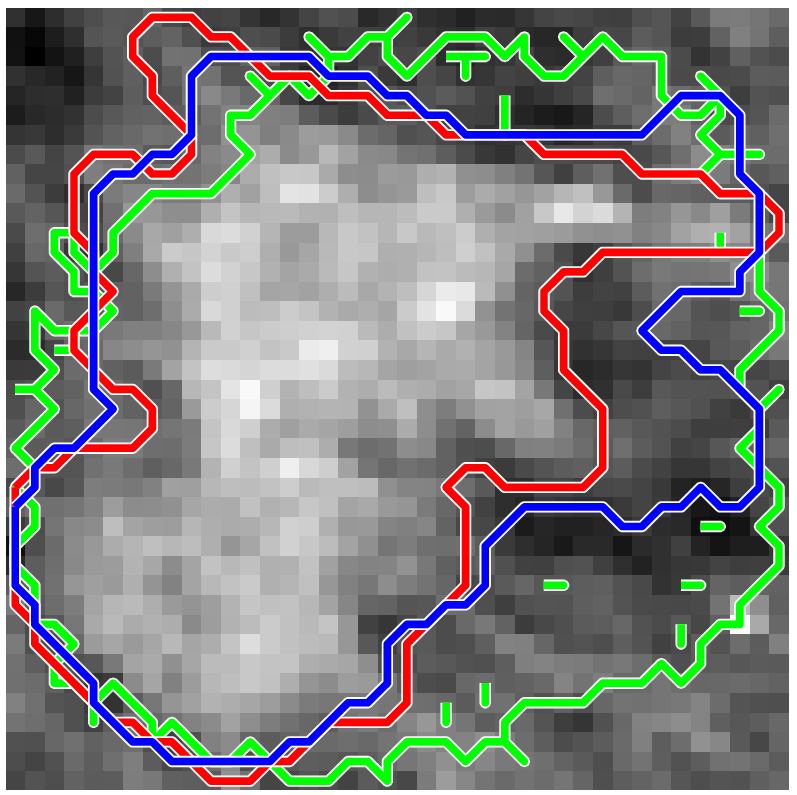}
		\caption{}
	\end{subfigure}
	\begin{subfigure}[b]{0.158\linewidth}
		\includegraphics[width=\linewidth]{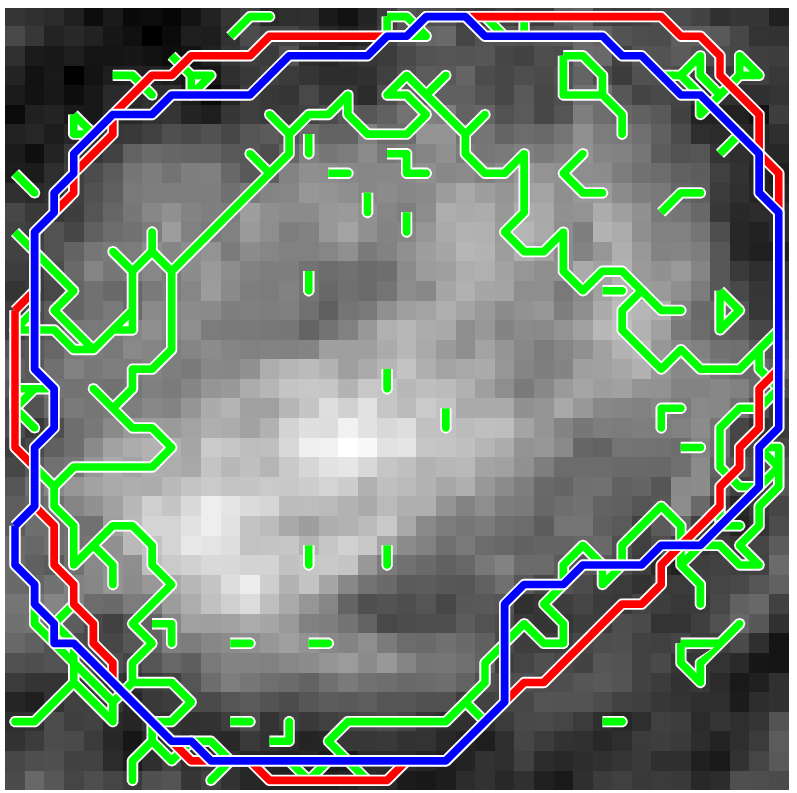}
		\caption{}
	\end{subfigure}
	\begin{subfigure}[b]{0.158\linewidth}
		\includegraphics[width=\linewidth]{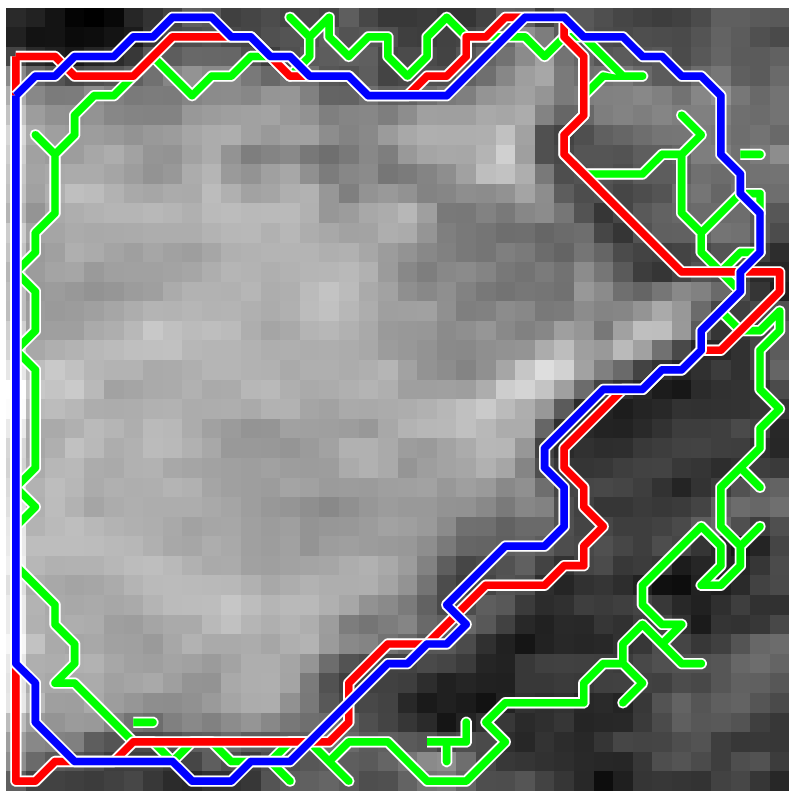}
		\caption{}
	\end{subfigure}
	\begin{subfigure}[b]{0.158\linewidth}
		\includegraphics[width=\linewidth]{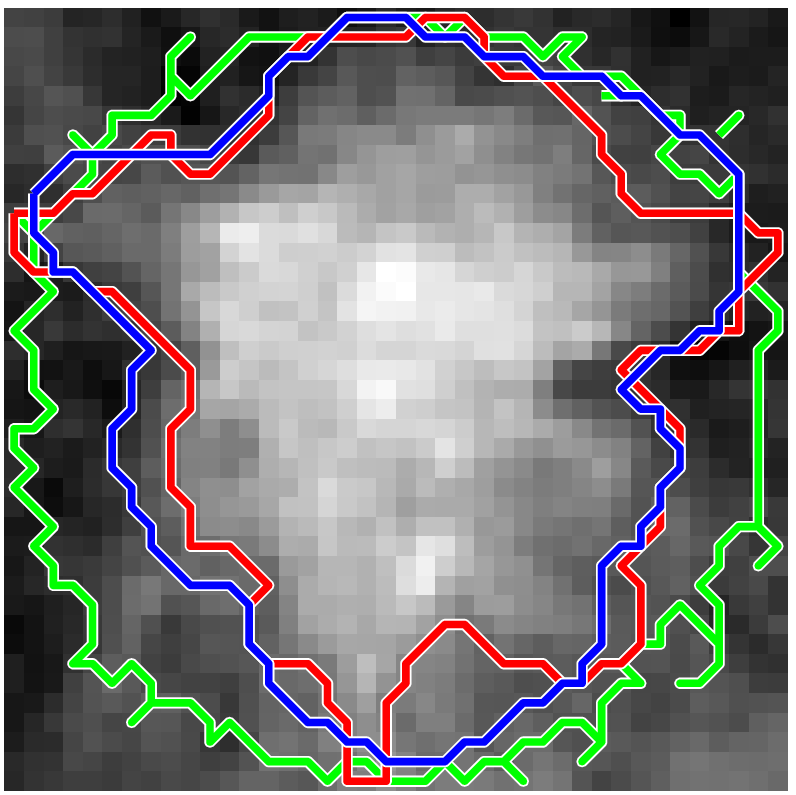}
		\caption{}
	\end{subfigure}
	\caption{Visualized comparison of segmentation results between CRU-Net ($\lambda=0.67$) method and Zhu's work on the 5 hardest cases, when Zhu's DI $\leq$ 81\%. Red lines denote the radiologist's contour, blue lines are the CRU-Net's results, and green lines are from Zhu's method. From (a) to (f), Zhu's DIs are: 70.16\%, 73.47\%, 76.11\%, 72.95\%, 80.36\% and 79.98\%. The CRU-Net's corresponding DIs are: 87.51\%, 92.43\%, 88.52\%, 95.01\%, 93.50\% and 91.33\%.}	
	\label{results_contour2}
\end{figure}

All state-of-art methods and the CRU-Net' performances are shown in the Table \ref{table_results}, where \cite{zhu2017adversarial} are reproduced, results of  \cite{beller2005example,cardoso2015closed,dhungel2015deep,dhungel2015deep1} are from their papers. Table \ref{table_results} shows that our proposed algorithm performs better than other published algorithms on both data sets. In INbreast, the best Dice Index (DI) 93.66\% is obtained with CRU-Net, No R ($\lambda=0.67$) and a similar DI 93.32\% is achieved by its residual learning; while in DDSM-BCRP, all state-of-art algorithm performs similarly and the best DI 91.43\% is obtained by CRU-Net ($\lambda=0$). The CRU-Net performs worse on DDSM-BCRP than INbreast, which is because of its worse data quality. To better understand the dice coefficients distribution in test sets, Fig. \ref{dice_pdfs} shows the histogram of dice coefficients and sampled cumulative distribution of two datasets. In those figures we can observe that the CRU-Net achieves a higher proportion of cases with DI $>95\%$. In addition, all algorithms follow a similar distribution, but Zhu's algorithm has a bigger tail than others on the INbreast data. To visually compare the performances, example contours from the CRU-Net ($\lambda=0.67$) and Zhu's algorithms are shown in Fig. \ref{results_contour}. It depicts that while achieving a similar DI value to Zhu's method, the CRU-Net obtains a less noisy boundary. To examine the tail in Zhu's DIs histogram (DI $\leq$ 81\%), Fig. \ref{results_contour2} compares the contours of the hard cases, which suggests that the proposed CRU-Net provides better contours for irregular shape masses with less noisy boundaries. 


\section{Conclusions}
In summary, we propose the CRU-Net to improve the standard U-Net segmentation performance via incorporating the advantages of probabilistic graphic models and deep residual learning. The CRU-Net algorithm does not require any tedious preprocesssing or postprocessing techniques. It outperforms published state-of-art methods on INbreast and DDSM-BCRP with best DIs as 93.66\% and 91.14\% respectively. In addition, it achieves higher segmentation accuracy when the applied database is of higher quality. The CRU-Net provides similar contour shapes (even for hard cases) to the radiologist with less noisy boundary, which plays a vital role in subsequent cancerous diagnosis.

%
% ---- Bibliography ----

\bibliographystyle{splncs04}
\bibliography{MassSegmentationRef}

\begin{thebibliography}{10}
\providecommand{\url}[1]{\texttt{#1}}
\providecommand{\urlprefix}{URL }
\providecommand{\doi}[1]{https://doi.org/#1}

\bibitem{ball2007digital}
Ball, J.E., Bruce, L.M.: Digital mammographic computer aided diagnosis (cad)
  using adaptive level set segmentation. In: Engineering in Medicine and
  Biology Society, 2007. EMBS 2007. 29th Annual International Conference of the
  IEEE. pp. 4973--4978. IEEE (2007)

\bibitem{beller2005example}
Beller, M., Stotzka, R., M{\"u}ller, T.O., Gemmeke, H.: An example-based system
  to support the segmentation of stellate lesions. In: Bildverarbeitung f{\"u}r
  die Medizin 2005, pp. 475--479. Springer (2005)

\bibitem{cardoso2015closed}
Cardoso, J.S., Domingues, I., Oliveira, H.P.: Closed shortest path in the
  original coordinates with an application to breast cancer. International
  Journal of Pattern Recognition and Artificial Intelligence  \textbf{29}(01),
  1555002 (2015)

\bibitem{carneiro2017review}
Carneiro, G., Zheng, Y., Xing, F., Yang, L.: Review of deep learning methods in
  mammography, cardiovascular, and microscopy image analysis. In: Deep Learning
  and Convolutional Neural Networks for Medical Image Computing, pp. 11--32.
  Springer (2017)

\bibitem{chen2017graph}
Chen, D., Lv, J., Yi, Z.: Graph regularized restricted boltzmann machine. IEEE
  transactions on neural networks and learning systems  \textbf{29}(6),
  2651--2659 (2018)

\bibitem{dhungel2015deep}
Dhungel, N., Carneiro, G., Bradley, A.P.: Deep learning and structured
  prediction for the segmentation of mass in mammograms. In: International
  Conference on Medical Image Computing and Computer-Assisted Intervention. pp.
  605--612. Springer (2015)

\bibitem{dhungel2015deep1}
Dhungel, N., Carneiro, G., Bradley, A.P.: Deep structured learning for mass
  segmentation from mammograms. In: Image Processing (ICIP), 2015 IEEE
  International Conference on. pp. 2950--2954. IEEE (2015)

\bibitem{he2016deep}
He, K., Zhang, X., Ren, S., Sun, J.: Deep residual learning for image
  recognition. In: Proceedings of the IEEE conference on computer vision and
  pattern recognition. pp. 770--778 (2016)

\bibitem{heath2000digital}
Heath, M., Bowyer, K., Kopans, D., Moore, R., Kegelmeyer, P.: The digital
  database for screening mammography. Digital mammography pp. 431--434 (2000)

\bibitem{pairwise_Gaussian}
Kr{\"a}henb{\"u}hl, P., Koltun, V.: Efficient inference in fully connected crfs
  with gaussian edge potentials. In: Advances in neural information processing
  systems. pp. 109--117 (2011)

\bibitem{moreira2012inbreast}
Moreira, I.C., Amaral, I., Domingues, I., Cardoso, A., Cardoso, M.J., Cardoso,
  J.S.: Inbreast: toward a full-field digital mammographic database. Academic
  radiology  \textbf{19}(2),  236--248 (2012)

\bibitem{oliver2010review}
Oliver, A., Freixenet, J., Marti, J., Perez, E., Pont, J., Denton, E.R.,
  Zwiggelaar, R.: A review of automatic mass detection and segmentation in
  mammographic images. Medical image analysis  \textbf{14}(2),  87--110 (2010)

\bibitem{ronneberger2015u}
Ronneberger, O., Fischer, P., Brox, T.: U-net: Convolutional networks for
  biomedical image segmentation. In: International Conference on Medical image
  computing and computer-assisted intervention. pp. 234--241. Springer (2015)

\bibitem{zheng2015conditional}
Zheng, S., Jayasumana, S., Romera-Paredes, B., Vineet, V., Su, Z., Du, D.,
  Huang, C., Torr, P.H.: Conditional random fields as recurrent neural
  networks. In: Proceedings of the IEEE International Conference on Computer
  Vision. pp. 1529--1537 (2015)

\bibitem{zhu2017adversarial}
Zhu, W., Xiang, X., Tran, T.D., Hager, G.D., Xie, X.: Adversarial deep
  structured nets for mass segmentation from mammograms. arXiv preprint
  arXiv:1710.09288  (2017)

\end{thebibliography}

\end{document}